\documentclass[letterpaper, 10 pt, conference]{ieeeconf}  

\IEEEoverridecommandlockouts                              

\usepackage{booktabs}
\usepackage{graphics} 
\usepackage{graphicx}
\usepackage{subfig}
\usepackage{multicol}
\usepackage{multirow}
\usepackage{amsmath} 
\usepackage{amssymb}  
\usepackage{hyperref}
\usepackage[nolist]{acronym}
\usepackage{cite}

\newcommand{\rbnote}[1]{\textcolor{green}{}}
\newcommand{\tsnote}[1]{\textcolor{red}{}}
\newcommand{\gpnote}[1]{\textcolor{blue}{}}

\DeclareMathOperator{\Exp}{Exp}
\DeclareMathOperator{\Log}{Log}

\title{\LARGE \bf
Riemannian Splat Regression Models for Learning Time Fields on Arbitrary Riemannian Manifolds}

\author{Anastasios Manganaris, Rohit Banerjee, Gokul Prabhakaran, and Ahmed H. Qureshi
\thanks{Anastasios Manganaris, Rohit Banerjee, Gokul Prabhakaran, and Ahmed H. Qureshi are with the Department of Computer Science, Purdue University, West
Lafayette, IN, USA, 47907. Email: \texttt{\{amangana, baner121, prabhag, ahqureshi\}@purdue.edu}}%
}

\begin{document}

\maketitle

\begin{figure*}[h!]
\centering
\subfloat[Torus $T^2$]{%
\includegraphics[height=5cm, keepaspectratio]{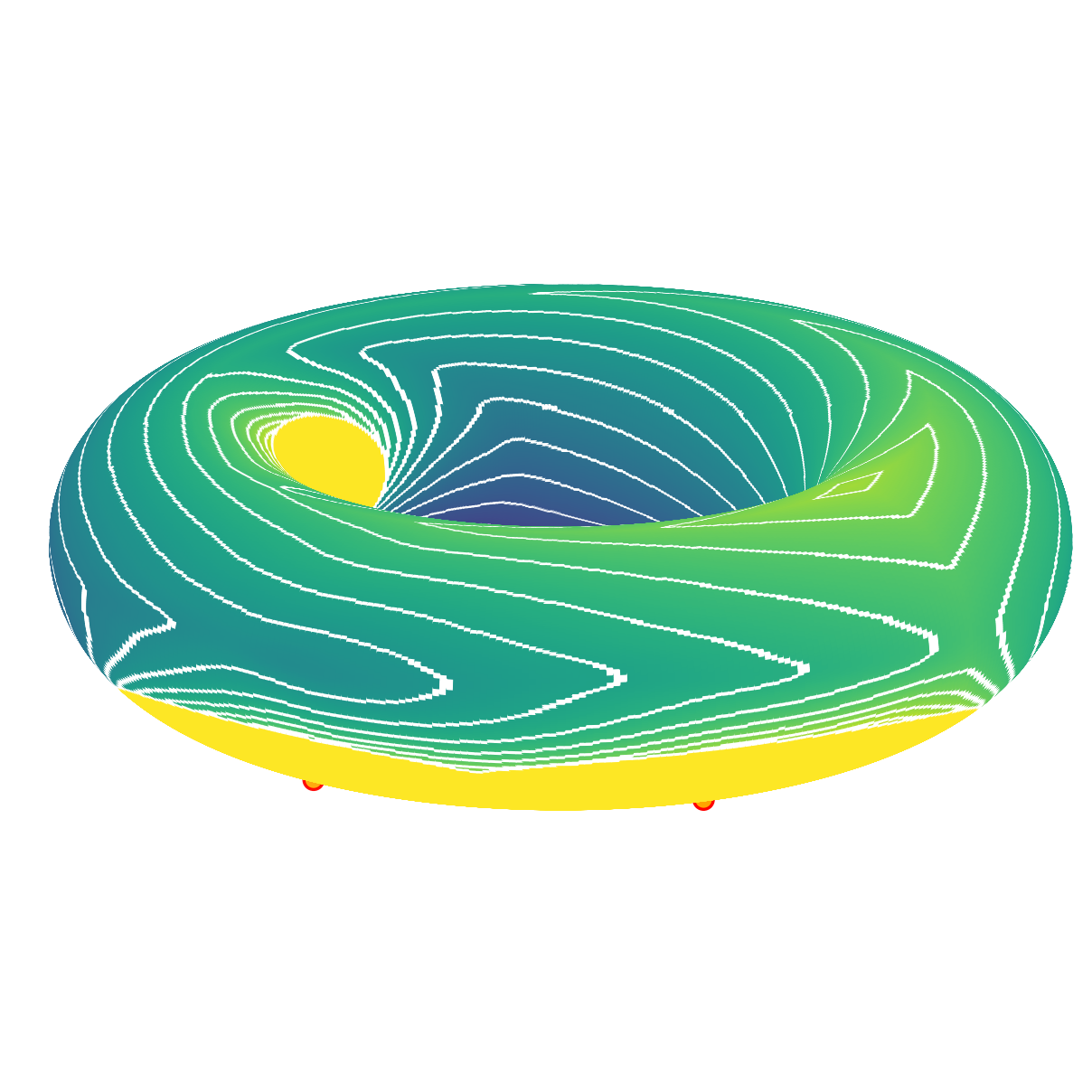}
\label{fig:intro:torus}
}
\hspace{0.005\linewidth}   
\subfloat[Sphere $S^2$]{%
\includegraphics[height=5cm, keepaspectratio]{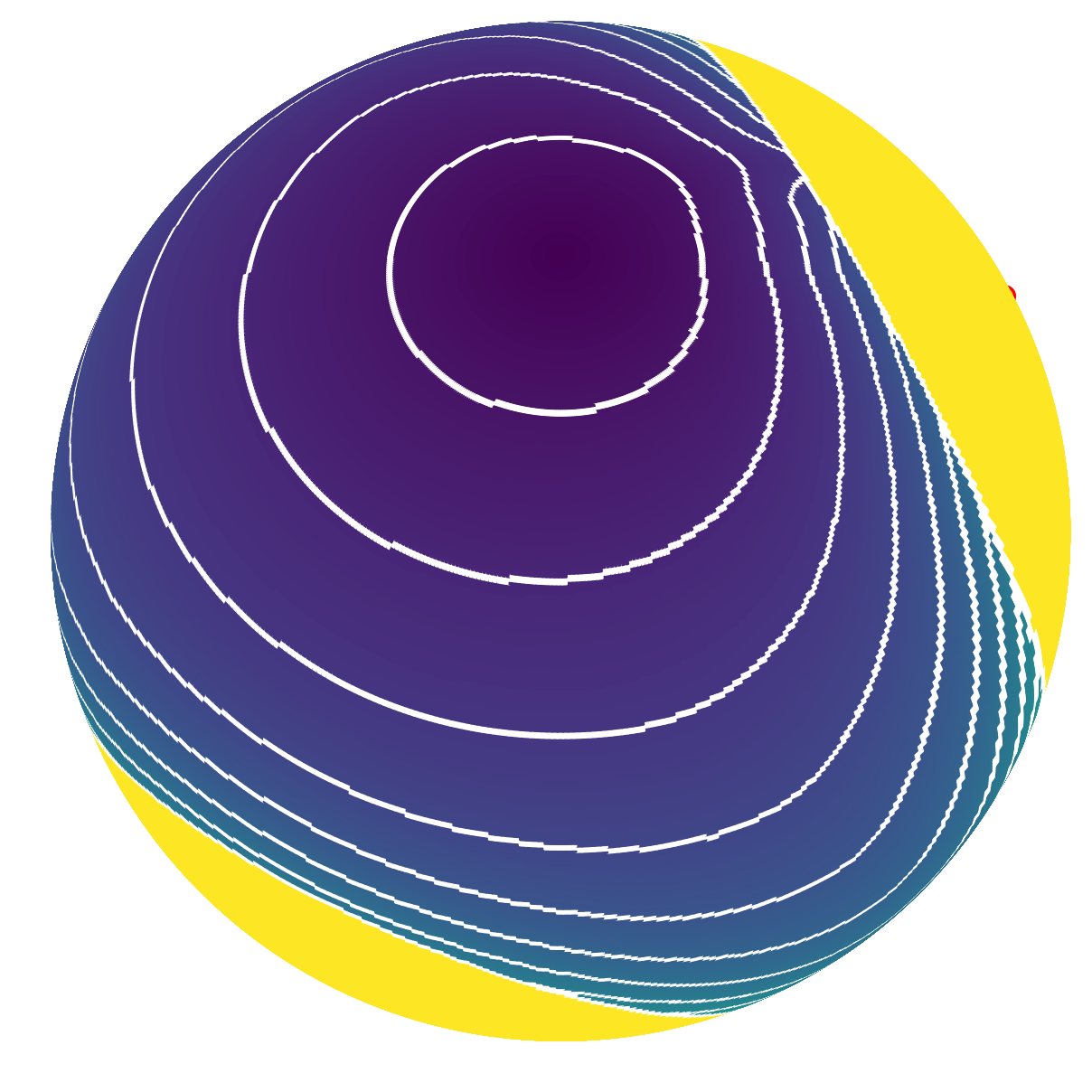}
\label{fig:intro:sphere}
}
\subfloat[(Lorentz) Hyperbolic $\mathcal{L}^2$]{%
\includegraphics[height=5cm, keepaspectratio]{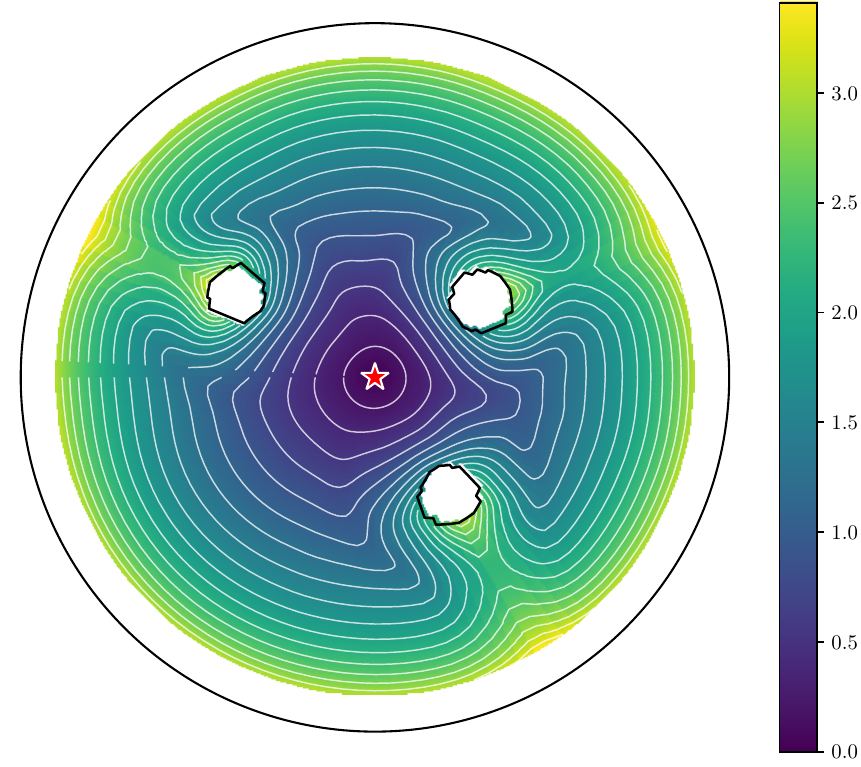}
\label{fig:intro:hyperbolic}
}
\caption{\acp{RSRM} can learn functions on the surface of differentiable manifolds such as torii, spheres, hyperbolic manifolds, and more without first requiring projecting coordinates into higher dimensional spaces and without using any model capacity on the off-manifold ambient space.}
\label{fig:intro}
\end{figure*}

\thispagestyle{empty}
\pagestyle{empty}

\begin{abstract}
Motion planning on arbitrary Riemannian manifolds is an important and difficult problem that frustrates typical planning methods for Euclidean spaces. In particular, motion planning methods that approximate optimal time-to-go functions with neural networks, e.g., Neural Time Fields (NTFields), cannot be directly applied without using ad-hoc coordinate projections into higher dimensions. Using these methods directly without such projections is desirable, as it promises to provide the lowest-possible-runtime method for obtaining optimal plans on high-dimensional manifolds while using minimal model capacity. In this work, we develop a model that requires no coordinate projection and can learn arbitrary functions on Riemannian manifolds by combining \acp{SRM} with splats defined by wrapped Gaussian distributions. We successfully apply this model for learning arrival time fields on several Riemannian manifolds, and we compare the accuracy and model size of this approach with \acp{MLP} adapted to work on each manifold individually.
\end{abstract}

\section{Introduction}

Motion planning for many robotic systems takes place in non-Euclidean spaces that feature global topological issues. 
For example, robot configuration spaces that include rotational components can be represented with the torus, sphere, and special orthogonal group manifolds. 
Another set of examples include configuration spaces with hierarchical structure, which can be embedded in hyperbolic manifolds~\cite{ManifoldBook-Ch3}.


There are existing motion planning methods that generally support such manifolds and optimal planning with respect to the metrics defined on them, with the most popular being sampling-based optimal planners \cite{Kingston_Zacha2018Sampling_Based}.
Our work pursues another category of motion planning methods based on solving a global optimal value function over the domain.
The main benefit of this approach is it provides the ability to query exact travel times and optimal paths between any two points at very fast speeds, at the cost of some pre-computation and memory usage \cite{Kimmel_Ron2004Fast_Marching_M, A_fast_marching_Sethia_1996}. 
Classical methods in this category computed this function over a discretized grid covering the domain, suffering from exponentially increasing computation time and space requirements as dimensionality increases.


More recent work in this category seeks to overcome this issue using more efficient function approximators \cite{ni2023ntfields}. 
While these have been successful in scaling such methods to much higher dimensions and supporting generalization over all start-goal-pairs, relatively little work has been invested in adopting them for learning value functions on arbitrary manifolds. 
For this, the main approach has just involved using coordinate projections to allow the neural network to learn a smooth function over a Euclidean domain that can be mapped back to the original manifold. 
However, this uses a larger-than-necessary amount of model capacity and doesn't generalize across manifolds. 
This prompts us to ask if there are other function approximators that could work for any Riemannian manifold.


Splat Regression Models \cite{daniels2026splat} are a recently proposed function approximator generalizing 3D Gaussian splatting~\cite{kerbl2023gaussiansplatting} for arbitary regression problems. 
These models are reported to be very efficient in comparison to other function approximators, especially in lower-dimensional regimes. Our main insight is that in combining \acp{SRM} with splats based on wrapped Gaussian distributions (being wrapped in the exponential map of an arbitrary Riemannian manifold) \cite{ManifoldBook-Ch4}, one obtains a function approximator for any Riemannian manifold. 
This paper contributes the formulation for this function approximator, and experiments demonstrating its usefulness for learning solutions to the Eikonal equation on several manifolds of interest to robotics. 
Our results show \acp{RSRM} to be capable function approximators that require no modification across manifolds and learn optimal travel time functions more efficiently than \acp{MLP}.

\section{Related work}

Our method is related to the broad domain of motion planning on manifolds. As mentioned, one of the most popular approaches for this task is based on extending sampling based motion planners to support constraints defined by some equality constraint $F(q) = 0$ on robot configurations $q \in Q$ \cite{Kingston_Zacha2018Sampling_Based}. These approaches technically handle a more general case than planning on differentiable manifolds; the space where $F(q) = 0$ is only a valid differentiable manifold when the Jacobian of $F$ is full-rank everywhere. These methods are broadly categorized by the way they enable sampling configurations under the constraints. These categories include relaxing constraints \cite{bonilla2015relaxing}, projecting sampled configurations onto the constraint manifold \cite{qureshi2022ConstrainedMPNets}, sampling nearly-on-manifold configurations using tangent-space steps from existing on-manifold points \cite{yakey2001tangentspace}, approximating the entire manifold using a precomputed atlas \cite{jaillet2013atlasrrt, Kim_Um_Suh_Park_2016}, and reparameterizing the manifold to enable closed form sampling over a space of parameters that map to manifold points \cite{mcmahon2014closedchain}.
Local planning can utilize similar techniques, such as projecting all local steps onto the manifold and approximately following the tangent space of the manifold at each point. Of course, however, this is not necessary for many manifolds of interest that support directly mapping from tangent space vectors at a point to a new point on the manifold through an exponential map or retraction.   
Finally, optimal sampling-based motion planners \cite{Sampling_based_Karama_2011, wilson2025aorrtc} can also support optimal planning with respect to a manifold's Riemannian metric. It is generally not possible to compute the geodesic distance between two arbitrary configurations for an arbitrary metric, so approximations are necessary, with existing work building approximations based on Gauss-Legendre quadratures \cite{din2026ritriemannianinformedtrees} or retractions \cite{kyaw2026geometryawaresamplingbasedmotionplanning}.

Another major category of methods does not resort to any approximation. They actually solve for the metric function over the manifold, usually with respect to a single or a few source points. For motion planning, the primarily relevant works are those that compute this function for all pairs of points on the manifold. After obtaining this function, it is possible to both very efficiently extract a geodesic path in the space and query the distance between points, which is useful in other decision making tasks related to vehicle routing and agent assignment. Methods compute this function as the solution of a particular \ac{PDE}. The particular \acp{PDE} used vary significantly. Earlier methods used \acp{PDE} describing the propagation of wavefronts, such as the Eikonal and HJB equations. Algorithms in this tradition include the Fast Marching Method \cite{A_fast_marching_Sethia_1996} and its extensions \cite{Ordered_upwind_Sethia_2001}, including those supporting non-Euclidian domains with arbitrary metrics \cite{mirebeau2019hamiltonianfmm}. More recent methods have been developed based on \acp{PDE} describing diffusion-processes, such as the heat equation \cite{crane2017heatmethod}. However, these have been applied primarily to single or few-source-geodesic-distance problems. We refer the reader to the survey by Crane et. al \cite{crane2020surveyalgorithmsgeodesicpaths} for additional information on these methods.

For robot motion planning, the above numerical grid-based methods are often insufficient due to their poor scaling in high dimensions, which is further exacerbated when attempting to compute all-pairs geodesic distance functions. However, the former category of \ac{PDE} based methods have recently given rise to methods for physics-informed motion planners that learn to approximate the all-pairs-geodesic-distance field mainly driven by a loss derived from the Eikonal equation \cite{ni2023ntfields}. These methods are significantly more scalable, successfully generalize over all pairs of points, and can still easily incorporate arbitrary Riemannian metrics by just adding a term in their training objective. We are interested in further extending these methods to efficiently learn functions over arbitrary differentiable manifolds. Due to using standard Euclidean function approximators like \acp{MLP}, ad-hoc coordinate projections are required to learn good approximations without seams or distortion. Our work explores a new function approximator class that works on any Riemannian manifold using minimal parameters and does not require any change of coordinates.

\section{Method}

We now describe our proposed function approximator, \acp{RSRM}, for learning optimal arrival-time fields for motion planning directly on curved configuration spaces.
These spaces are defined as Riemannian manifolds $(\mathcal{M}, g)$, where $\mathcal{M}$ is a differentiable manifold and the Riemannian metric $g$ determines local lengths, angles and gradient norms.
For background, we first review \acp{SRM}~\cite{daniels2026splat}, which express functions on Euclidian domains as a sum of many weighted Gaussian basis functions.
We then review wrapped Gaussian distributions~\cite{ManifoldBook-Ch4}, which serve as an alternative basis function that is locally projected onto the manifold through its exponential map.
Finally, we combine the above to obtain \acp{RSRM} for approximating arrival-time field, trained via enforcing the Eikonal equation at sampled points on the manifold. 

\subsection{Splat Regression Models}

An \ac{SRM}~\cite{daniels2026splat} represents a real-valued function as a weighted sum of localized basis functions, called \emph{splats}.
Let $\rho:\mathbb{R}^d\to\mathbb{R}_{\geq 0}$ denote the standard Gaussian density function.
The simplest \ac{SRM} places $k$ copies of $\rho$ at centers $\mu_i \in \mathbb{R}^d$: 
\begin{equation}
    f(x) = \sum_{i=1}^{k} v_i \, \rho(x - \mu_i),
    \label{eq:srm-simple}
\end{equation}
where $v_i \in \mathbb{R}$ is the weight of the $i^{th}$ splat.
Since a Gaussian rapidly decays away from its center, each splat affects the function primarily in a local neighborhood of $\mu_i$. The full model, given by
\begin{equation}
    f(x) = \sum_{i=1}^{k} v_i \, \rho\!\left(A_i^{-1}(x - \mu_i)\right) \cdot |\det A_i^{-1}|,
    \label{eq:srm}
\end{equation}
associates an invertible matrix $A_i \in \mathbb{R}^{d\times d}$
with each splat, allowing them to have a different sizes and orientations. 
Due to this simple structure, \acp{SRM} permit capacity to be concentrated where the field is difficult to approximate and even dynamically adjusted during training by simply adding or removing splats. We utilize this adaptive capacity in our implementation.




\subsection{Wrapped Gaussian Splats}

The subtraction $x-\mu_i$ in (\ref{eq:srm}) is defined only in Euclidean vector space. 
Points on a general manifold cannot be subtracted, so we replace this operation to obtain a similar displacement vector for the manifold.
For each $\mu \in \mathcal{M}$, the tangent space $T_\mu\mathcal{M}$ is a Euclidean vector space that locally approximates the manifold near $\mu$.
The exponential map, $\Exp_\mu:T_\mu\mathcal{M}\rightarrow\mathcal{M}$, maps a tangent vector to the endpoint of the geodesic that begins at $\mu$ with that initial direction and length. 
On a neighborhood in which $\Exp_\mu$ is injective, its local inverse is the logarithmic map $\Log_\mu:\mathcal{M}\rightarrow T_\mu\mathcal{M}$, which maps a nearby point $x$ to the tangent vector at $\mu$ pointing toward $x$.
This logarithmic map provides the needed notion of on-manifold displacement from the splat center $\mu$.
Combined with the an invertible matrix $A \in \mathbb{R}^{d\times d}$ as in \eqref{eq:srm}, the density given by a \emph{wrapped} Gaussian splat at $x\in\mathcal{M}$ is
\begin{equation}
    \begin{split}
    \rho_W(x;\mu,A) = \rho\!\left(A^{-1}\Log_\mu(x)\right) & \cdot \left|\det D\Log_\mu(x)\right| \\
    \cdot \left|\det A^{-1} \right|,
    \end{split}
    \label{eq:wrapped-gaussian}
\end{equation}
where $D\Log_\mu(x)$ is the Jacobian of the logarithmic map at $x$. 
This determinant term corrects for the local change in volume induced by the map between the manifold and its tangent space~\cite{ManifoldBook-Ch4}. 


\subsection{Riemannian Splat Regression Models}

Equipped with the definition of $\rho_W$, \acp{RSRM} can be simply evaluated at a point by summing the individual densities from each splat at that point, i.e.,
\begin{equation}
    f(x) = \sum_{i=1}^{k} v_i\,\rho_W(x; \mu_i, A_i).
    \label{eq:rsrm}
\end{equation}
For each splat located at $\mu_i$, the input point is passed through $\Log_{\mu_i}$ and then through $A_i^{-1}$ to obtain a normalized tangent space point. The standard Gaussian density is computed at that point and then corrected by the absolute determinant of the Jacobian of $\Log_{\mu_i}$ and $A_i^{-1}$. The individual contributions are then weighted and summed to obtain the final value. Training an \ac{RSRM} updates $v_i$, $\mu_i$, and $A_i$ for all splats, and only differs from training an \ac{SRM} in that the splat centers $\mu_i$ are constrained to remain on the manifold over which the function is defined. This is simply achieved by projecting the points back onto the manifold after each parameter update.

\subsection{Learning Arrival Times}
\label{sec:arrival_times}

We use \acp{RSRM} to learn the optimal arrival-time field for a fixed goal configuration $x_g \in \mathcal{M}$. 
Let $T(x)$ denote the minimum time required to travel from an arbitrary configuration $x$ to $x_g$. 
Motion is specified by a slowness field (reciprocal of speed) $s: \mathcal{M}\to[1,\infty)$, where $s(x)$ is the time required per unit Riemannian distance at $x$.
Obstacles and undesirable regions are assigned large slowness values, whereas free space has slowness near a value of one. 
The arrival-time field satisfies the Eikonal equation:
\begin{equation}
    \|\nabla T(x)\|_g=s(x), \qquad T(x_g)=0,
    \label{eq:eikonal}
\end{equation}
where $\nabla T(x)$ is the Riemannian gradient of $T$, and $\|\cdot\|_g$ is the norm induced by the Riemannian metric $g$. 
Following the negative gradient of a valid arrival-time field produces a path that locally decreases arrival time and leads toward the source. We parameterize the arrival-time field as 
\begin{equation}
    T(x) = d_{\mathcal{M}}(x,x_g) / f(x),
    \label{eq:factored}
\end{equation}
where $d_{\mathcal{M}}(x,x_g)$ is the geodesic distance from any point $x$ to $x_g$ and $f$ is the RSRM in \eqref{eq:rsrm}. 
This construction imposes the source condition exactly, since $d_{\mathcal{M}}(x_g,x_g)=0$. 

We train the model without precomputed arrival-time labels. 
At each optimization step, we sample collocation points $x$ from a distribution $q$ over $\mathcal{M}$ and minimize the squared Eikonal residual,
\begin{equation}
    \mathcal{L}_{\mathrm{Eik}} = \mathbb{E}_{x\sim q}
    \left[\left(\|\nabla T(x)\|_g-s(x)\right)^2\right]
    \label{eq:eikonal-loss}
\end{equation}
The prescribed slowness field supplies the supervision, while automatic differentiation evaluates the gradient of the parameterized field. After training, paths are obtained by following the negative Riemannian gradient of $T$ from a query configuration toward $x_g$.

\section{Results}

\begin{figure}[t]
    \centering
    \includegraphics[width=1.0\linewidth]{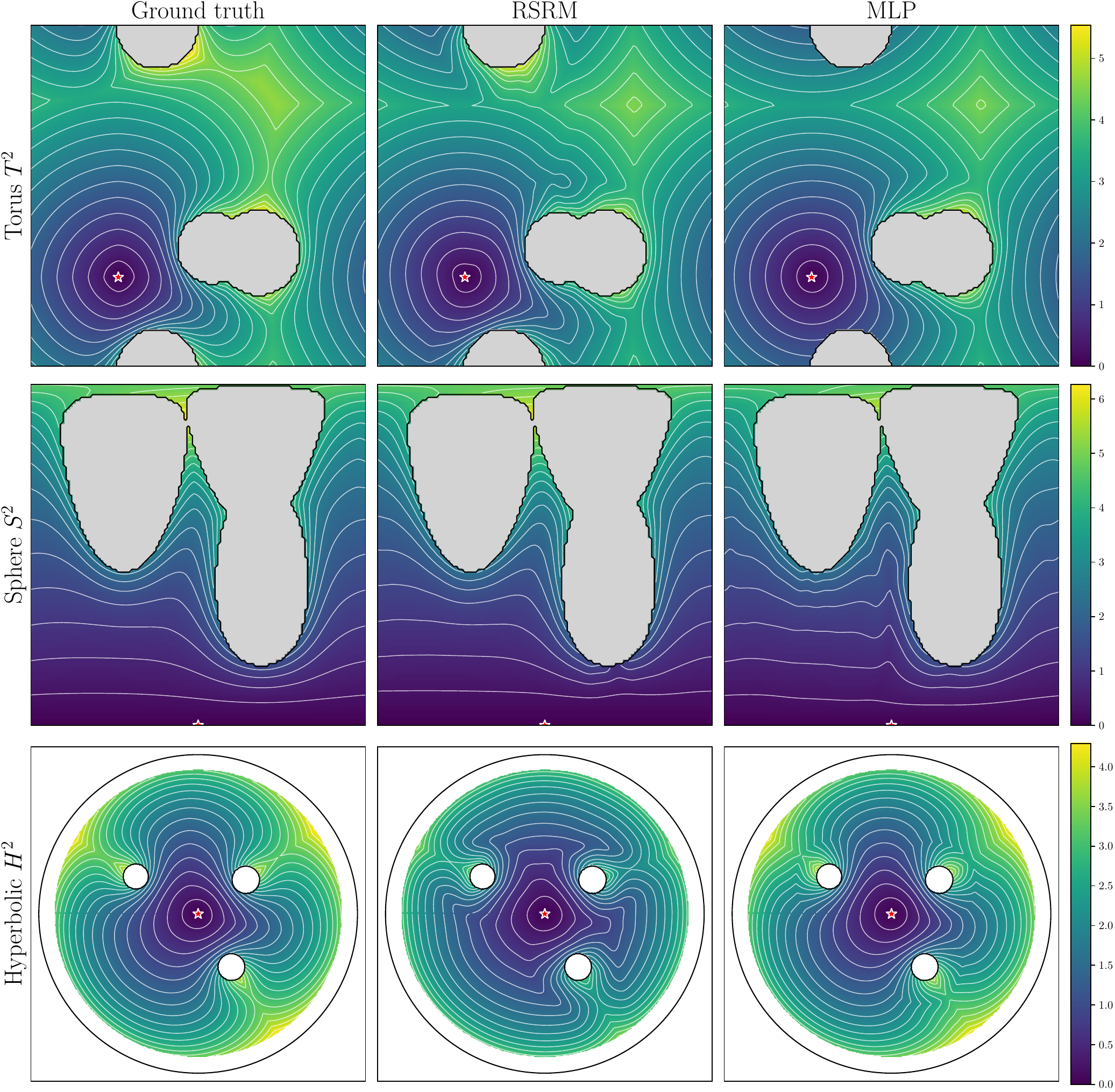}
    \caption{\acp{RSRM} are capable of learning arrival time fields on torus, sphere, and hyperbolic manifolds without modification.}
    \label{fig:results}
\end{figure}

We evaluated \acp{RSRM} in comparison to \acp{MLP} for learning arrival time fields on three ubiquitous manifolds: a torus, sphere, and hyperbolic space. We trained both in five randomized environments per manifold with equal training and collocation point budgets. Our results, visualized in Figure~\ref{fig:results} and summarized by Table~\ref{tab:results}, demonstrate \acp{RSRM} are capable of learning accurate arrival time fields on these manifolds. In comparison to standard \ac{MLP} architectures trained the same way but individually adjusted to respect the structure of each manifold, we observe \acp{RSRM} are generally more accurate while using far fewer parameters.

The two-dimensional torus first demonstrates \acp{RSRM}' natural support for periodic domains. The manifold has coordinates in $[-\pi, \pi]^2$ and uses the Logarithmic map $\Log_{\mu}(x_j) = (x_j + \pi) \bmod (2\pi) - \pi$ for each coordinate $j$. As shown in Fig.~\ref{fig:results}, the \ac{RSRM} learns an accurate arrival time field that smoothly propagates through the domain's boundaries. The \ac{MLP} baseline, which requires a sin-cos embedding of the input coordinates to avoid seams at those boundaries, has slightly lower error while using many more parameters. The two-dimensional sphere embedded in $R^3$ demonstrates how the \ac{RSRM} can use splats defined in two-dimensional tangent spaces for a manifold embedded in a three-dimensional space. For this curved manifold, the \ac{RSRM} is slightly more accurate while again using significantly fewer parameters than the \ac{MLP}.
For both the sphere and the torus, the total training time is greater with the \ac{SRM} than with the \ac{MLP}. The majority of this time is taken in periodically enlarging and recompiling the \ac{SRM}, as the times taken for each optimization step are nearly equal at this model size.

The two-dimensional hyperbolic manifold using the Lorentz model is similarly embedded in $R^3$ but is not periodic like the sphere or torus. For this domain, it is necessary to remove the sin-cos embedding layer of the \ac{MLP} and significantly increase its size to obtain reasonable accuracy, while no such modification is required for the \ac{RSRM}. While the \ac{RSRM} does not obtain as low error on this manifold, it still achieves a mean error slightly better than the \ac{MLP} using only 3\% of the \ac{MLP}'s parameters. Moreover, the training time for the \ac{SRM} is also less than that required for the \ac{MLP} in this domain, due to the additional time-per-step needed for larger \acp{MLP}.

\begin{table}[t]
\centering
    \caption{Comparison of our approach with \ac{MLP} on three manifolds. Mean and standard deviation are reported across five training seeds.}
    \label{tab:results}
    \begin{tabular}{llcc}
        \toprule
        & & \textbf{RSRM (Ours)} & \textbf{MLP} \\
        \midrule
        \multirow{3}{*}{Final Loss} & Torus & $\boldsymbol{0.0236 \pm 0.0048}$ & $0.106 \pm 0.0677$ \\
         & Sphere & $\boldsymbol{0.0517 \pm 0.0171}$ & $0.111 \pm 0.0394$ \\
         & Hyperbolic & $\boldsymbol{0.0238 \pm 0.0042}$ & $0.0333 \pm 0.0343$ \\
        \midrule
        \multirow{3}{12mm}{Time-per-Step (ms)} & Torus & $1.69 \pm 0.04$ & $\boldsymbol{1.61 \pm 0.01}$ \\
         & Sphere & $4.74 \pm 0.08$ & $\boldsymbol{3.22 \pm 0.03}$ \\
         & Hyperbolic & $\boldsymbol{7.61 \pm 0.33}$ & $23.68 \pm 0.06$ \\
        \midrule
        \multirow{3}{10mm}{Training \\ Time (s)} & Torus & $61.5 \pm 1.7$ & $\boldsymbol{28.9 \pm 2.2}$ \\
         & Sphere & $124.6 \pm 4.9$ & $\boldsymbol{33.5 \pm 0.7}$ \\
         & Hyperbolic & $\boldsymbol{257.0 \pm 68.3}$ & $257.9 \pm 9.6$ \\
        \midrule
        \multirow{3}{*}{Final Error} & Torus & $\boldsymbol{0.276 \pm 0.108}$ & $0.324 \pm 0.139$ \\
         & Sphere & $\boldsymbol{0.095 \pm 0.0502}$ & $0.152 \pm 0.0342$ \\
         & Hyperbolic & $\boldsymbol{0.995 \pm 0.393}$ & $1.03 \pm 0.6$ \\
        \midrule
        \multirow{3}{*}{Parameters} & Torus & $\boldsymbol{3534 \pm 18}$ & $33793 \pm 0$ \\
         & Sphere & $\boldsymbol{3854 \pm 153}$ & $34049 \pm 0$ \\
         & Hyperbolic & $\boldsymbol{6531 \pm 1584}$ & $198657 \pm 0$ \\
        \bottomrule
    \end{tabular}
\end{table}
  
\section{Conclusion}

We introduced \acp{RSRM}, a modification to \acp{SRM} using wrapped Gaussian distributions to enable function approximation on arbitrary Riemmanian manifolds. We evaluated \acp{RSRM} in learning the optimal arrival time fields on three ubiquitous manifolds and found them to be capable of learning these functions accurately based only on self-supervision from the Eikonal equation, while using far fewer parameters than an \ac{MLP} baseline. These results, in addition to their qualititative benefits (manifold-agnostic design, local editability, and flexible model size) are compelling evidence for using \acp{RSRM} for on-manifold function approximation and arrival-time-field-based motion planning. Future work will investigate further increasing accuracy of learned arrival-time functions and applying the approach to more complex environments in higher dimensions. 

\bibliographystyle{IEEEtran}
\bibliography{IEEEabrv,references}

\begin{acronym}[TDMA]
\acro{SRM}{Splat Regression Model}
\acro{RSRM}{Riemannian Splat Regression Model}
\acro{PDE}{Partial Differential Equation}
\acro{MLP}{Multi-layer Perceptron}
\end{acronym}

\end{document}